%% file: main.tex
\documentclass[letterpaper]{article} 
\usepackage{aaai24}  
\usepackage{times}  
\usepackage{helvet}  
\usepackage{courier}  
\usepackage[hyphens]{url}  
\usepackage{graphicx} 
\usepackage{natbib}  
\usepackage{caption} 
\usepackage{algorithm}
\usepackage{algorithmic}

\usepackage{newfloat}
\usepackage{listings}
\DeclareCaptionStyle{ruled}{
  labelfont=normalfont,
  labelsep=colon,
  strut=off,
} 
\floatstyle{ruled}
\newfloat{listing}{tb}{lst}{}
\floatname{listing}{Listing}

\usepackage[
  capitalise,
  noabbrev,
]{cleveref}
\usepackage{subcaption}

\title{
  Maintaining User Trust Through Multistage Uncertainty Aware Inference
}
\author {
  Chandan Agrawal,\textsuperscript{\rm 1}
  Ashish Papanai,\textsuperscript{\rm 1}
  Jerome White
}
\affiliations{
  \textsuperscript{\rm 1}Wadhwani Institute for Artificial Intelliegence\\
  agri-ai@wadhwaniai.org
}

\begin{document}

\maketitle

\begin{abstract}
  \input{abstract}
\end{abstract}

\input{intro}
\input{related}
\input{setup}
\input{method}
\input{results}

\input{conclusion}
\input{acknowledgments}

\bibliography{main.bib}
\end{document}

%% file: abstract.tex
This paper describes and evaluates a multistage approach to AI
deployment. Each stage involves a more accurate method of inference,
yet engaging each comes with an increasing cost. In outlining the
architecture, we present a method for quantifying model uncertainty
that facilitates confident deferral decisions.
The architecture is currently under active deployment to thousands of
cotton farmers across India. The broader idea however is applicable to
a growing sector of AI deployments in challenging low resources
settings.

%% file: intro.tex
\section{Introduction}
\label{sec:intro}

Artificial intelligence is playing an increasingly prominent role in
decision support systems. This change has opened a new line of work in
building systems that users can trust and on which they can safely
rely~\cite{trust-review}. This paper considers a scenario in which
users have justifiably high standards for model performance, however
deployment conditions make using state-of-the-art modeling techniques
difficult. Our solution is to use a cascade of models that balance
response time with recommendation accuracy.

For the last several years our group has developed and deployed
computer vision models to help small holder cotton farmers make pest
management decisions~\cite{kdd-exposed}. A mobile app first asks
users---farmers and farming extension workers---to take photos of
bollworms caught in pest traps around their field. The images are then
passed to the vision model, trained to identify and count bollworms in
the photo. Counts are used to provide pesticide spraying advice based
on rules from reputed agricultural bodies.

One of our challenges has been maintaining user trust. Like many users
of AI systems~\cite{expectations}, our audience has high expectations
of model performance. This makes it easy to damage the relationship we
have with users when incorrect recommendations are suggested. In many
cases that means decreased chances of repeat usage, as farmers have
little time for tools that are not reliable. In more extreme cases the
fallout is worse: panic that a crop is seriously endangered, and
allocation of limited resources toward unnecessary spraying.

There are a number of ways to address this deployment challenge
responsibly. This paper presents the technical approach we have taken:
a pipeline of increasingly accurate, decreasingly accessible, models
that defer to one another based on their evaluation of uncertainty. A
variant the models presented here, and their encompassing
architecture, is currently in active deployment to thousands of cotton
farmers across India.

%% file: related.tex
\section{Related Work}
\label{sec:related}

Work in human-AI decision systems has introduced several methods for
increasing trust among users~\cite{human-ai-review}. A popular
technique for doing so is incorporating model uncertainty into the
workflow. Some methods directly expose users to model
confidence~\cite{uncertainty-1, uncertainty-2}. The idea being that
such transparency increases user agency. This methodology puts an onus
on interface design and usability education, something that is outside
the scope of our work.

Other efforts use uncertainty to suppress model output. Some propose
outright rejection~\cite{ml-rejection}, while others defer to based on
models of human expertise~\citet{pmlr-defer}. \citet{wilder-ijcai}
optimize ``human-machine teamwork'' by training models that take into
account the strengths each for a given task. This as opposed to a
model that is fixed during human evaluation time. They report strong
improvements in this joint setup. As with our work,
\citet{ml-rejection} use box confidence to determine uncertainty,
however they do not do it in the context of multistage modeling.

There is evidence that slowing down the inference process improves
user interaction~\cite{park-slow}. Our architecture inherently does
that in some cases.

%% file: setup.tex
\input{figures/flow}

\section{Problem Setup}
\label{sec:setup}

\subsection{Pest detection in low resource settings}
\label{sec:setting}

Our users are cotton farmers and farming extension officers who have
placed specialized pest traps throughout their fields. The traps are
designed to capture pink and American bollworms, which have been a
menace to Indian cotton farmers for decades~\cite{pbw-return}. The
key to mitigating crop loss caused be these pests is early detection,
something the specialized traps facilitate.

Under normal operation, the traps are periodically emptied and their
contents analyzed. The number of bollworms caught is an indicator of
what action to take. Our app allows this process to be supported by
AI: photos of the trap contents are fed to a model that is trained to
identify pink and American bollworms; rule-based logic within the app
turns those counts into action recommendations.

While there have been several contributions to AI-based pest
detection~\cite{agri-review-1, agri-review-2}, successful deployment
poses challenges beyond modeling. Because our app is used in rural
settings, we cannot rely on consistent mobile internet
connectivity. In addition, users are sensitive to app size footprints,
being hesitant to download specialty apps that are more-than about
50MB. To meet these requirements, our primary in-app model must be
small and capable of operating offline.

\subsection{Exposing optimal inference}

In general, smaller models are not as performant as those with more
parameters. Our own baseline experiments confirm that generality in
this context. Because of the importance users place on our estimates,
it is prudent to explore all means of providing optimal model
performance.

The motivation for our approach came from observations from our own
on-ground studies. First, while users prefer instant feedback---spray
recommendations that come during natural app interaction---they are
not necessarily wedded to that. In small scale experiments, we found
users willing to wait up to 24~hours for a result. Second, rural
internet connectivity is not always binary. There are several cases in
which connectivity is fluid throughout the field, and others in which
it occurs long after a field visit. Such cases can arise either
because of tower-environment dynamics, or because of travel away from
the field after trap monitoring.

\subsection{A multistage modeling approach}

To take advantage of these observations, we propose a multistage
architecture that sequentially passes the image to larger models,
culminating with a human-in-the-loop (\cref{fig:flow}). The first
model in the sequence resides on the phone, while the second model
lives in the cloud. The third stage in the sequence is human
judgment. Which model estimate is presented to the user depends on the
uncertainty surrounding that models judgment. The final stage of the
pipeline involving humans is assumed to be certain and accurate.

%% file: figures/flow.tex
\begin{figure*}[t]
  \centering
  \includegraphics{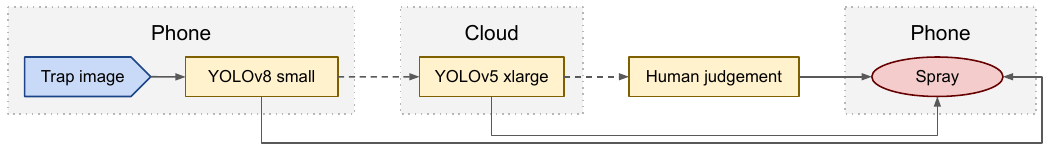}
  \caption{Proposed multistage architecture. Images are first passed
    to a small model on the phone. If that model is uncertain about
    its predictions, the images are sent to a larger model in
    cloud. If the cloud model is uncertain, a human expert is
    engaged. If the phone or cloud model is confident in its estimate,
    the pipeline is halted and a recommendation is made.}
  \label{fig:flow}
\end{figure*}

%% file: method.tex
\section{Methodology}
\label{sec:method}

\input{figures/thresholds}

\subsection{Object detection for count classification}

The objective of the system is to provide a spray recommendation given
an image of trapped bollworms. Recommendations take the form of a
three-valued alert depending on the number of bollworms present: no
action required; imminent spray required; or ``be cautious,'' akin
something in between. Within this paper we use a generally accepted
count-to-alert mapping as follows: zero pests represents no action
required, one to seven pests be cautious, and eight or more spray. We
train and evaluate our models using an open source dataset designed
for bollworm model development~\cite{bollwm}.\footnote{Data set
version 20220912-2056.} Evaluation in this paper was performed over
the validation set of 2093 images: 698 no action, 728 cautious, 667
spray.

Our phone and cloud models (\cref{fig:flow}) are based on object
detection systems. This means that the systems are not trained to
classify, but to focus on pest recognition. Counts from the
recognition are turned into alerts using the rules described
above. Both models are variants of the YOLO family~\cite{yolo}, and
within that based on derivatives from Ultralytics. The phone model is
YOLOv8~\emph{small}, while the cloud model is
YOLOv5x6~\emph{xlarge}. The models consist of 11.2~million and
43.7~million parameters respectively. The relatively low parameter
count of small allows it to fit within our mobile app footprint
budget.

By default, YOLO models produce a large number of boxes at different
levels of confidence. A majority of the boxes are often invalid, with
confidence values very close to zero. A standard model culls these
low-confidence estimates via a parameter passed at inference. Our
method eschews this parameter, considering all boxes. It empirically
explores windows of confidence values over which model estimates can
be trusted. Specifically a lower and upper bound that when applied at
inference proves robust to proper classification of alert
thresholds. Where the model is unsure, it abstains rather than
estimates.

\subsection{Box confidence windowing}

Consider a collection of boxes and corresponding box confidences
produced by YOLO inference of an image. Box confidences range from
zero to one, exclusive. A \emph{window} consists of an upper and lower
value that partitions the collection into two sets. The first are
boxes with confidence greater-than the lower value; the second boxes
with confidence greater than upper. The lower set acts to cull noise,
while the upper focuses on high confidence.

Uncertainty is determined by calculating the difference in cardinality
between the two sets, relative the alert boundaries described
earlier. Let $l$ and $u$ represent the cardinalities of the first
(lower established) and second (upper established) partitions,
respectively. If $l$ and $u$ produce the same alert, the model
estimate is accepted and a recommendation made. If that is not the
case, the model abstains, sending the message to the next evaluator in
the chain (\cref{fig:flow}).

We determine the ideal window empirically, by studying tendencies over
the validation set. \cref{fig:thresholds} displays such tendencies of
the phone model. Model performance is measured using Matthews
correlation coefficient (MCC). Performance tends to be extreme---very
good or very bad, relatively---at high abstention levels. This occurs
when the lower threshold is zero, and when the upper threshold is
above about 0.85. As we move away from those areas---toward the
diagonal, for example---there a number of potentially acceptable
windows that maximize performance while minimizing abstention.

Controlling the balance between performance and abstention is
important. While better classification is the goal, abstention has
both a usability and an operational cost. Phone abstention forces
users to wait longer for a recommendation. For example, in a recent
deployment the phone model took well under a second to respond. The
mean cloud response time however was approximately seven hours. The
mode of the cloud distribution was even higher (almost 12 hours)
suggesting highly variable internet access across our users. When the
cloud also abstains, human experts must make time to review the
images. A cloud model that rarely abstains is ideal.

%% file: figures/thresholds.tex
\begin{figure*}[t]
  \centering
  \begin{subfigure}[b]{0.49\textwidth}
    \centering
    \includegraphics[width=\textwidth]{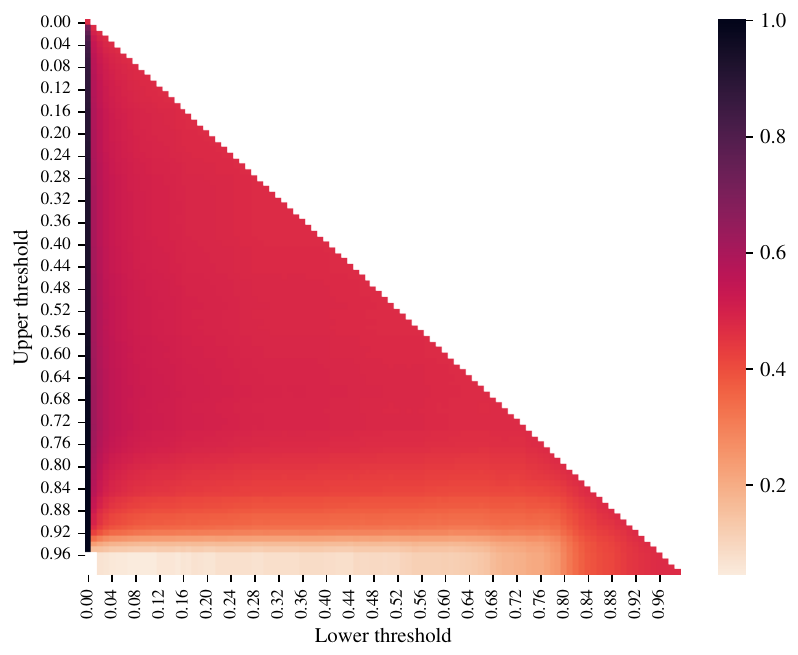}
    \caption{Relationship between thresholds and MCC.}
    \label{fig:thresholds-performance}
  \end{subfigure}
  \hfill
  \begin{subfigure}[b]{0.49\textwidth}
    \centering
    \includegraphics[width=\textwidth]{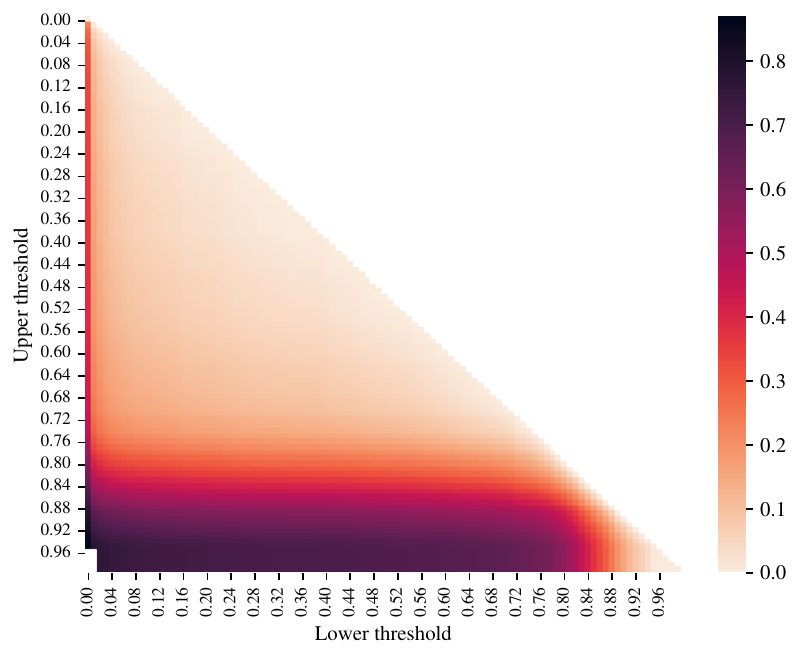}
    \caption{Relationship between thresholds and fraction of abstentions.}
    \label{fig:thresholds-abstention}
  \end{subfigure}
  \caption{Visualization of the dynamics between the lower and upper
    box confidence thresholds in the phone model.}
  \label{fig:thresholds}
\end{figure*}

%% file: results.tex
\section{Results}
\label{sec:results}

This section examines \emph{combined} models that operate
sequentially, as outlined in \cref{fig:flow}. These models show an
improvement to the phone model. They also offer a framework by which
we can engage stakeholders to make more informed deployment decisions.

\subsection{End-to-end performance}

\cref{fig:results} displays the performance (MCC) of our
\emph{combined} model. The $x$-axis represents the abstention level of
the phone model; the $y$-axis that of the cloud model ``conditioned''
on the phone. A conditioned system is developed as follows:
let a model \emph{candidate} be the output of a model at a given
confidence window. Phone candidates, for example, are cells in the
heatmaps of \cref{fig:thresholds}. We first group phone candidates by
their abstention fraction, then select a single model within each. In
cases where there are multiple candidates for a given fraction, the
candidate with the highest MCC is selected. For each candidate, the
process is repeated for the cloud model. However, only images on which
the phone candidate abstained is the respective cloud model
evaluated. This is what is meant by conditioning. Images over which a
cloud candidate abstains are assumed to be human-judged and given
predictions that match the ground truth. This process provides a
realistic picture of how the models will perform in practice.

\cref{fig:results} shows that as the phone model abstention fraction
increases, so to does performance. The lower-left portion of the
figure represents the system if only the phone were in operation. The
upper-right represents performance if only human intervention was
used. By examining cells in other regions, we can make informed
deployment decisions that balance model performance with user and
operational staff burden.

\input{figures/results}

\subsection{Comparative performance}
\label{sec:results-compare}

While \cref{fig:results} is informative, it does not give a sense for
how much more beneficial the combined model is as compared to other
types of single model deployments. \cref{fig:comparison} is an effort
to bridge that gap. The curves display the fraction of \emph{false
alarms}~(FA) made across abstention fractions. A false alarm is a case
in which the model erroneously makes a spray recommendation. As
discussed in the introduction, such cases can be expensive and
distressing for our users.

It is difficult to compare models directly because of the difference
in images at similar abstention fractions, and because in the combined
model there are several candidates from which to choose. To provide a
fair evaluation framework, we first group our phone candidates by
abstention fraction. For each candidate a cloud candidate is selected
that has the same inclusion set. The candidate with the lowest~FA
score is taken in the case of ties. All candidate
models are taken from the combined model set, however only abstained
images are excluded before the candidate is evaluated. The result of
this process is shown in \cref{fig:comparison}.

Performance is highly variable across abstention levels; and because
the combined model is composed of several candidates at each, it
is highly variable within an abstention fraction. For this reason,
curves have been smoothed using a sliding window. Windows use median
aggregation to reduce strong expert biases in combined model sets.

Despite this processing---and the sacrifices that come with making
consistent evaluation sets at each abstention level---the curves
provide a good sense for how models compare. The combined model
presented in this paper consistently outperforms the phone model, and
even the cloud model in some cases. The performance difference is
especially notable at low abstention fractions (less-than 0.2). This
is encouraging because such fractions represent acceptable levels for
our deployment.

\input{figures/comparison}

%% file: figures/results.tex
\begin{figure}[t]
  \centering
  \includegraphics[width=\columnwidth]{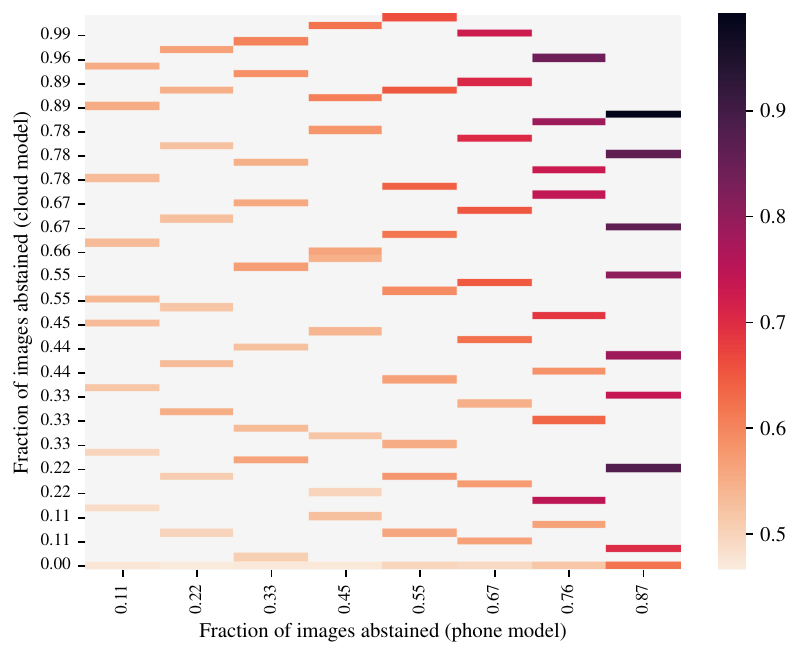}
  \caption{Performance of the end-to-end system at varying abstention
    levels. Cells represent MCC for that phone-cloud abstention
    combination.}
  \label{fig:results}
\end{figure}

%% file: figures/comparison.tex
\begin{figure}[t]
  \centering
  \includegraphics[width=\columnwidth]{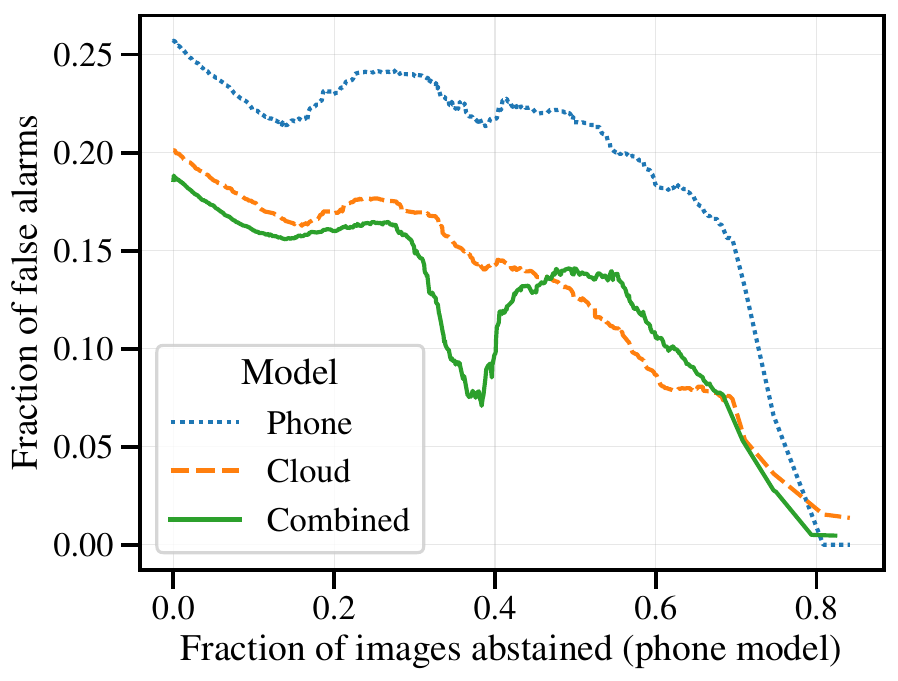}
  \caption{Fraction of erroneous spray recommendations at varying
    abstention levels. ``Combined'' represents our end-to-end system.}
  \label{fig:comparison}
\end{figure}

%% file: conclusion.tex
\subsection{Conclusion and Future Work}

This paper presents a variant of the architecture and methodology we
have used to deploy a decision support system to farmers across
India. The use of a staged model to increase trust and reliability is
a novel improvement over our previous
efforts~\cite{kdd-exposed}. Framing the modeling effort in a duality
between performance and response time has been helpful in engaging our
partners and users.

Immediate future work includes evaluating our on-ground
performance. At the time of writing, we are nearing the end of the
primary cotton growing season in India, and in turn our active
deployment. A deeper understanding of our frameworks performance is
forthcoming.

The approach taken in this paper has largely been empirical. Framing
the problem as one of optimization will allow us to reason about the
proposed approach more robustly; picking windows deterministically, or
taking response time into account, for example. It will also allow us
to build and compare other approaches. In our own experiments,
treating the problem as one of classification rather than threshold
based object detection has yet to show promise. However it is a worthy
effort that deserves exploration.

Finally, picking thresholds based on training set abstention rates is
limiting. Ideally, we should also take downstream resource
availability into account. For example, the phone model optimizing for
network availability, or the cloud model accounting for the human
expert limitations. Such an approach has
precedence~\cite{sumedh-nature, pmlr-defer}, and is a line of work
worth pursuing.

Responsible AI deployment is an imperative; especially as AI systems
become ingrained in increasingly more critical tasks. Using a
multi-stage approach is one we are excited to see become more broadly
adopted.

%% file: acknowledgments.tex
\section*{Acknowledgements}

We thank our fellow team members for their various contributions to
the project: Jatin Agarwal, Ayushi Bhotica, Soma Dhavala, Sonali
Ghike, Fahad Khan, Abhishek Kumar, Anika Murarka, Srinivasa Rao,
Mohammad Salman, and JP Tripathi. We also thank the anonymous
reviewers for their insightful feedback.